\documentclass[conference,9pt]{IEEEtran}

\usepackage{todonotes}
\usepackage{amsmath}
\usepackage{cite}
\usepackage{makecell}
\usepackage{bm}
\usepackage{multirow}
\usepackage{makecell}

\usepackage{color}

\newcommand{\mytilde}{\raise.17ex\hbox{$\scriptstyle\sim$}}
\usepackage{threeparttable}

\makeatletter
\newcommand*\titleheader[1]{\gdef\@titleheader{#1}}
\AtBeginDocument{%
  \let\st@red@title\@title
  \def\@title{%
    \bgroup\normalfont\normalsize\centering\@titleheader\par\egroup
    \vskip1ex\st@red@title}
}
\makeatother

\title{Control Variate Approximation for DNN Accelerators}
\titleheader{\vspace{-12pt}To appear at the 58th Design Automation Conference (DAC'21), December 5-9, 2021, San Francisco, USA.}

\begin{document}

\author{\IEEEauthorblockN{Georgios Zervakis\IEEEauthorrefmark{1},
Ourania Spantidi\IEEEauthorrefmark{2},
Iraklis Anagnostopoulos\IEEEauthorrefmark{2},
Hussam Amrouch\IEEEauthorrefmark{3},
and J{\"o}rg Henkel\IEEEauthorrefmark{1}
}
\IEEEauthorblockA{\IEEEauthorrefmark{1}Chair for Embedded Systems (CES), Karlsruhe Institute of Technology, Karlsruhe, Germany}
\IEEEauthorblockA{\IEEEauthorrefmark{2}Department of Electrical, Computer and Biomedical Engineering, Southern Illinois University, Carbondale, U.S.A.}
\IEEEauthorblockA{\IEEEauthorrefmark{3}Chair of Semiconductor Test and Reliability (STAR), University of Stuttgart, Stuttgart, Germany}
\IEEEauthorblockA{
\IEEEauthorrefmark{1}\{georgios.zervakis,henkel\}@kit.edu,
\IEEEauthorrefmark{2}\{ourania.spantidi,iraklis.anagno\}@siu.edu,
\IEEEauthorrefmark{3}amrouch@iti.uni-stuttgart.de
}
\vspace{-15pt}
}

\maketitle
\begin{abstract}
In this work, we introduce a control variate approximation technique for low error approximate Deep Neural Network (DNN) accelerators.
The control variate technique is used in Monte Carlo methods to achieve variance reduction.
Our approach significantly decreases the induced error due to approximate multiplications in DNN inference, without requiring time-exhaustive retraining compared to state-of-the-art.
Leveraging our control variate method, we use highly approximated multipliers to generate power-optimized DNN accelerators.
Our experimental evaluation on six DNNs, for Cifar-10 and Cifar-100 datasets, demonstrates that, compared to the accurate design, our control variate approximation achieves same performance and 24\% power reduction for a merely 0.16\% accuracy loss.
\end{abstract}
\vspace{1ex}


\section{Introduction}

Recent Deep Neural Networks (DNNs) have brought advancements in many fields over the past years.
This wide variety of fields includes various applications, such as computer vision, speech recognition, scientific computing and many more~\cite{jouppi2017datacenter}.
However, the late DNN advancements have brought up the immense demands for computational power
as well as for energy efficiency.
This need is intensified especially in deployments on smart and IoT edge devices, since they have very restricted energy and computing resources.
In addition, the increased DNN workload has led to the emerge of customized hardware DNN accelerators~\cite{jouppi2017datacenter}.

The main arithmetic computation during inference is the multiply-accumulate (MAC) operation.
DNNs perform millions of MAC operations on their convolution and fully-connected layers.
Therefore, DNN accelerators integrate thousands of MAC units.
For example, Google TPU~\cite{jouppi2017datacenter} comprises 64K MAC units while Google Edge TPU contains a 4K MAC units.
This vast number of MAC units combined with high parallelization leads to very high power demands~\cite{amrouch2020npu}.

Recently, approximate computing emerged as a promising solution to develop energy/power-efficient circuits.
Approximate computing leverages the intrinsic error resilience of a vast number of application domains to improve their energy profile at the cost of some accuracy loss~\cite{SaadatTCAD2018}.
Since DNNs are characterized by error-resilience, they inherently become appropriate candidates for approximate computations~\cite{SaadatTCAD2018,sarwar2018energy}.
Driven by this high potential for energy efficiency, significant research interest is shown in the design of approximate DNN accelerators~\cite{SaadatTCAD2018,sarwar2018energy,mrazek2016design,tasoulas2020weight,alwann,zervakis2020design,hanif2019cann}.
State of the art mainly approximates the multipliers of the DNN accelerator, since the multiplier is the most power consuming component of the MAC unit.
Modern DNNs are becoming gradually deeper, ending up with multiple layers that can differ significantly in error resilience~\cite{alwann}.
In addition, the errors due to approximate circuits are not constant but they are highly input dependent~\cite{zervakis2020design}.
Hence, different layers within the neural network or different neural networks require different approximation to satisfy an accuracy threshold~\cite{tasoulas2020weight,alwann,zervakis2020design}.
Moreover, recent research showed that the deeper the neural network, the more sensitive it becomes to any approximation~\cite{tasoulas2020weight}.
To overcome the aforementioned limitations, existing works mainly apply retraining to mitigate the accuracy loss due to approximation~\cite{mrazek2016design,sarwar2018energy}.
However, for DNNs, retraining targeting a specific approximate accelerator is very time consuming.
In addition, in many cases the training set might not be available (e.g., proprietary models) and thus, retraining might be infeasible~\cite{alwann}.

In this work, we introduce a \textit{control variate approximation technique} that improves the accuracy of approximate DNN accelerators without the need to perform time-overwhelming retraining.
Leveraging the accuracy improvement of our control variate approximation, we use~\cite{ZervakisTVLSI2016} to replace the multipliers of the DNN accelerator with aggressive approximate ones (i.e., with high error but also with high power gain).
The approximate multiplier of~\cite{ZervakisTVLSI2016} is selected because i) it delivers very high power reduction, albeit its high error and ii) its error can be analytically modeled with simple equations.
Our analysis demonstrates that our technique nullifies the mean value of the convolution error, and decreases its variance.
Our experimental evaluation, over six DNNs trained on Cifar-10 and Cifar-100, shows that, for a merely $0.16$\% average accuracy loss, our approach delivers $24$\% power reduction and same performance compared with the accurate design.
Our control variate approximation technique can be applied with any approximate multiplier as long as it is based on mathematical formulation instead of custom of logic simplification.

\textbf{Our novel contributions within this paper are as follows:}\\
(1) We propose a \textit{control variate approximation technique} for generating approximate DNN accelerators. Our technique does not require re-training and delivers high accuracy by controlling the error induced by the approximate multiplications.\\
(2) We demonstrate that our technique improves the accuracy up to $21$\%, on average, compared to exactly the same approximate DNN accelerator, but without our proposed control variate approximation. \\
(3) Leveraging our proposed control variate approximation, we can integrate highly approximate multipliers, thus significantly reduce the power consumption (more than $24$\%) of the whole DNN accelerator.
In addition, our technique outperforms state-of-the-art works, achieving more than $3.8$x higher energy reduction.

\section{Related works}
There has been great interest around approximate computing for neural network inference.
\cite{mrazek2016design} employed approximate multipliers to different convolution layers and~\cite{sarwar2018energy} proposed a compact and energy-efficient multiplier-less artificial neuron.
However,~\cite{mrazek2016design,sarwar2018energy} are based on retraining to recover accuracy loss caused by the usage of approximation.
Similar to~\cite{mrazek2016design}, \cite{hanif2019cann} utilized approximate multipliers and introduce an error compensation module.
However,~\cite{mrazek2016design,hanif2019cann} are evaluated on the LeNet network, a very shallow architecture which cannot provide the amount of operations recent DNNs do.
Therefore, both of these methods can be deemed inapplicable in modern scenarios which require deeper network architectures. 
The authors in~\cite{alwann} propose a non-uniform architecture that utilized approximate multipliers from~\cite{mrazek2017evoapproxsb}.
Their work tunes the weights accordingly and avoids retraining.
\cite{alwann} applies layer-wise approximation, while power-gating the unused approximate multipliers.
In~\cite{zervakis2020design}, approximate multipliers with reconfigurable accuracy at run-time are generated.
Similar to~\cite{alwann}, they also apply layer-wise approximation, which is narrowing down potential benefits.
In~\cite{tasoulas2020weight}, approximate reconfigurable multipliers with low variance are generated using~\cite{zervakis2020design}.
Then,~\cite{tasoulas2020weight} proposed a mapping algorithm to implement weight-oriented approximate inference in which the accuracy level of the approximate multiplier is determined by the weight's value.
Nevertheless, to enable runtime reconfiguration and accuracy control,~\cite{zervakis2020design,tasoulas2020weight} deliver limited energy savings due to the additional hardware.

\textbf{Distinguish from Existing State of the Art:} Our technique controls the error of the approximate multiplications and improves the accuracy without requiring re-training.
Consequently, this high accuracy improvement enables the exploitation of highly approximated multipliers, maximizing the power gains.

\section{Control Variate Approximation} \label{sec:error}
This section describes our control variate approximation technique and presents an error analysis of the approximate convolution.
The core operation of a convolution is given by:
\begin{small}\begin{equation}\label{eq:conv}
G=B+\sum_{j=1}^{k}{W_j\cdot A_j},
\end{equation}\end{small}\noindent
where B is the bias of the neuron, $W_j$ are the weights, and $A_j$ are the input activations.
In our approximate architecture and error analysis, approximate multipliers are used to replace the accurate multipliers of the DNN accelerator.
We denote $\epsilon_j$ the multiplication error of the product $W_j\cdot A_j$.
By error, we refer to the difference between the accurate and the approximate results.
Thus, $\epsilon_j = W_j\cdot A_j -  W_j\cdot A_j|_{approx}$.
Given \eqref{eq:conv}, the convolution error, namely $\epsilon_G$, is equal to:
\begin{small}
\begin{equation}\label{eq:converror}
\epsilon_G = B+\sum_{j=1}^{k}{W_j\cdot A_j}-B-\sum_{j=1}^{k}{W_j\cdot A_j|_{approx}} 
           = \sum_{j=1}^{k}{\epsilon_j}.
\end{equation}
\end{small}

The error value of an approximate multiplier can be considered as a random variable, and is therefore defined by its mean value and variance~\cite{LiVAR3}.
Denoting the mean and variance of the approximate multiplier by $\mu_{AM}$ and $\sigma^2_{AM}$, the mean and variance of the convolution operation are given by:
\begin{small}
\begin{equation}\label{eq:meanvar}
\begin{split}
\mathrm{E}[\epsilon_G] &=\mathrm{E}\big[\sum_{j=1}^{k}{\epsilon_j}\big]=k\mu_{AM}\\
\mathrm{Var}(\epsilon_G) &=\mathrm{Var}\Big(\sum_{j=1}^{k}{\epsilon_j}\Big)=k\sigma^2_{AM}.
\end{split}
\end{equation}
\end{small}\noindent
Note that, the error values $\epsilon_j$ are independent variables and thus their covariance is zero~\cite{LiVAR3,tasoulas2020weight}.

Hence, even if the approximate multiplier features small error (small $\mu_{AM}$ and $\sigma^2_{AM}$), the convolution error is significantly higher since it is proportional to the filter's size as \eqref{eq:meanvar} demonstrates.
In~\cite{tasoulas2020weight}, approximate multipliers with systematic error are employed and a constant correction term is used to compensate for the mean error (i.e., $\mathrm{E}[\epsilon_G]$).
However, even in this case, \textit{the error of the convolution is still high, since it is defined by its high variance ($Var(\epsilon_G)$).}

In our work, we propose the utilization of a control variate technique to reduce the convolution error variance.
A control variate is an easily evaluated random variable, with known mean, that is highly correlated with our variable of interest.
To implement our control variate approximation technique, instead of~\eqref{eq:conv}, we compute~\eqref{eq:axconv}.
\begin{small}
\begin{equation}\label{eq:axconv}
G^\ast=B+\sum_{j=1}^{k}{W_j\cdot A_j|_{approx}}+V
\end{equation}
\end{small}
\indent{}For the approximate multiplication ($W_j\cdot A_j|_{approx}$), we use the state-of-the-art power efficient perforated multipliers~\cite{ZervakisTVLSI2016}.
In~\cite{ZervakisTVLSI2016}, the $m$ least partial products are perforated, i.e., they are not generated, and thus, they are removed from the accumulation tree.
The latter results to smaller tree architectures that exhibit high delay, area, and power gains~\cite{ZervakisTVLSI2016}.
Moreover, a significant feature of~\cite{ZervakisTVLSI2016} is that it applies functional approximation.
Hence, its error does not depend on any carry propagation and it can be modeled in a mathematically rigorous manner.
Nevertheless, our control variate approximation and our analysis hereafter can be employed with any approximate multiplier that exhibits a similar behavior with~\cite{ZervakisTVLSI2016}, i.e., the induced error can be described by an analytical model.

When perforating the $m$ least partial products, the multiplication error of~\cite{ZervakisTVLSI2016} is obtained by:
\begin{small}
\begin{equation}\label{eq:pperr}
\begin{split}
\epsilon_j &= W_j\cdot A_j -  W_j\cdot A_j|_{approx} \\
                   &= W_j\cdot A_j - W_j\cdot (A_j-A_j\text{ mod }2^m) \\
                   &= W_j\cdot x_j, \text{ with } x_j = A_j\text{ mod }2^m = A_j\ \&\ (2^m-1)
\end{split}
\end{equation} 
\end{small}\noindent
Note that, for each filter in the DNN, the values of the weights ($W_j$) are fixed and are known a priori (after training and quantization).
Thus, only the $x_j$ change at run-time.
In addition, $x_j \in [0, 2^m-1]$ and thus requires only $m$-bits.

Considering the multiplication error $\epsilon_j$ in~\eqref{eq:pperr}, we set our control variate $V$ equal to:
\begin{small}
\begin{equation}\label{eq:v}
V=\sum_{j=1}^{k}{v_j}=\sum_{j=1}^{k}{x_j\cdot C_j},\ v_j=x_j\cdot C_j 
\end{equation}
\end{small}\noindent
where $C_j$ is a constant.
Hence, $\epsilon_j$ and $V_j$ feature a perfect linear correlation.
Obviously, selecting $C_j=W_j$ will deliver accurate results.
However, calculating $x_j\cdot W_j$ is computationally expensive and neglects the gains (area, power) of the perforated multiplier, since calculating $V$ requires $k$ multiplications and $k-1$ additions.
Note that a control variate is easily evaluated.
For this reason, we set $C_j=C$, $\forall C_j$.
As a result, $V$ is given by:
\begin{small}
\begin{equation}\label{eq:v2}
V=C\cdot\sum_{j=1}^{k}{x_j} \text{ and }  v_j=x_j\cdot C.
\end{equation}
\end{small}\noindent
Note that to calculate $V$ in~\eqref{eq:v2}, only $k-1$ additions and $1$ multiplication are required.

Given~\eqref{eq:v2}, the approximate convolution~\eqref{eq:axconv} is then written as:
\begin{small}
\begin{equation}\label{eq:axconv2}
\begin{split}
G^\ast &=B+\sum_{j=1}^{k}{\Big(W_j\cdot A_j-\epsilon_j\Big)} + \sum_{j=1}^{k}{v_j}\\
       &=G - \sum_{j=1}^{k}{\Big(\epsilon_j - v_j \Big)}
\end{split}
\end{equation}
\end{small}\noindent
and thus, the approximate convolution error equals:
\begin{small}
\begin{equation}\label{eq:axconverr}
\epsilon_{G^\ast}=\sum_{j=1}^{k}{\Big(\epsilon_j - v_j \Big)}
                 =\sum_{j=1}^{k}{\Big(x_j\cdot(W_j-C) \Big)}.
\end{equation} 
\end{small}

\begin{figure}[t!]
\centering
\resizebox{0.85\columnwidth}{!}{\includegraphics{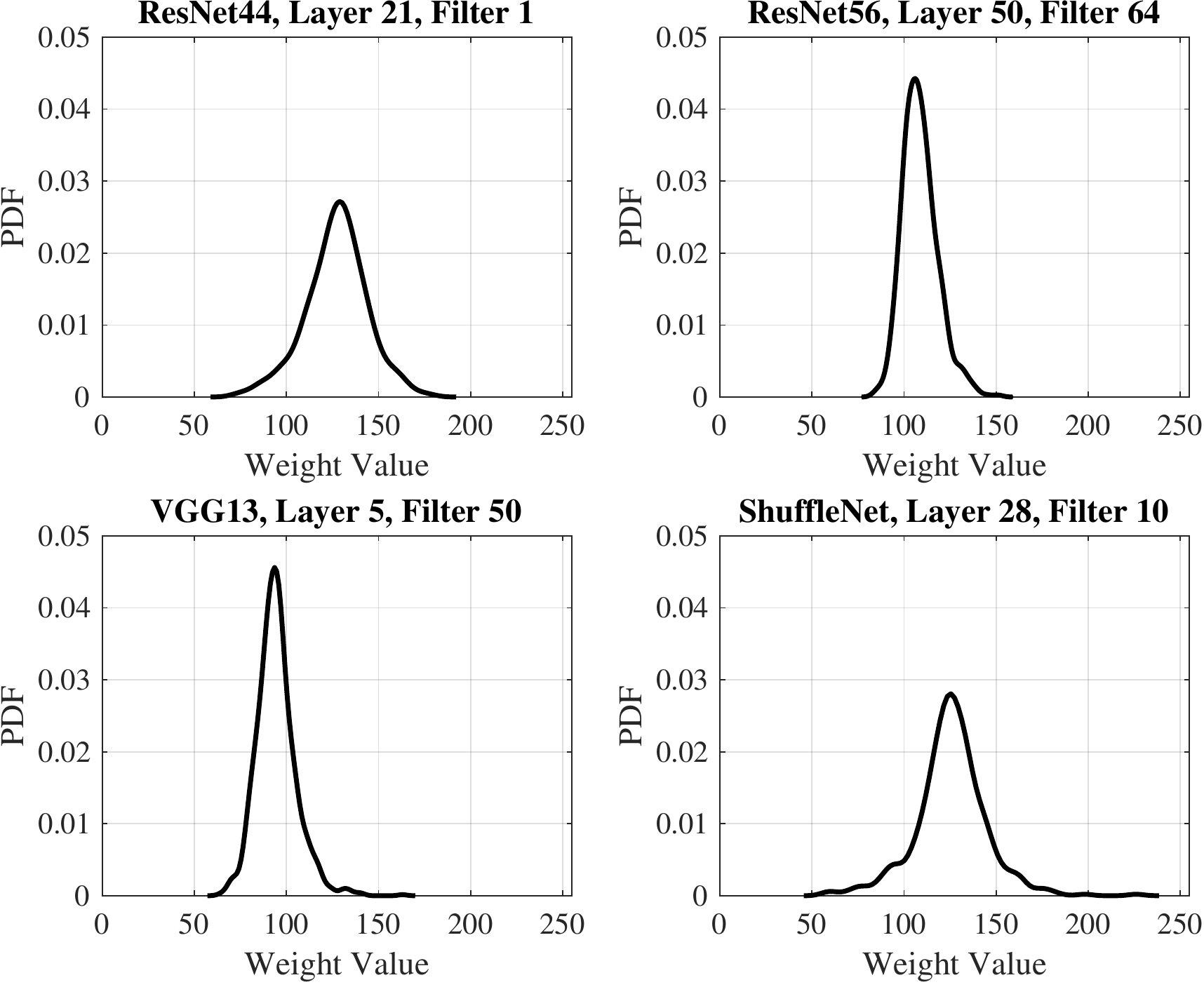}}
\caption{Weight distribution of randomly selected filters of various NNs. Four examples are depicted.}
\label{fig:Wdistr}
\vspace{-4pt}
\end{figure}

Hence, the variance $\mathrm{Var}(\epsilon_{G^\ast})$ of the error of the approximate convolution, i.e., $\epsilon_{G^\ast}$, is calculated from:
\begin{small}
\begin{equation}\label{eq:var}
\begin{split}
\mathrm{Var}(\epsilon_{G^\ast}) & = \sum_{j=1}^{k}{\mathrm{Var}\Big(\epsilon_j - v_j \Big)} \\
                       & = \sum_{j=1}^{k}{\Big((W_j-C)^2\cdot \mathrm{Var}(x_j)\Big)} \\
                       & = \underbrace{\frac{(2^m-1)(2^m+1)}{12}}_{\mathrm{Var}(x_j)}\sum_{j=1}^{k}{(W_j-C)^2}.
\end{split}
\end{equation} 
\end{small}\noindent
As a result, $\mathrm{Var}(\epsilon_{G^\ast})$ is minimized when:
\begin{small}
\begin{equation}\label{eq:minvar}
\begin{gathered}
\frac{d}{dC}\mathrm{Var}(\epsilon_{G^\ast}) = 0 \Rightarrow \\
C=\mathrm{E}[W_j]=\frac{1}{k}\sum_{j=1}^{k}{W_j}.
\end{gathered}
\end{equation} 
\end{small}\noindent
Proofs of~\eqref{eq:var} and~\eqref{eq:minvar} are simple and they are omitted due to space limitation.
Note that $C\neq 0$, i.e., variance without our control variate (as in~\eqref{eq:meanvar}).
In addition, note that the more squeezed the weights' distributions is (i.e., concentrated close to $\mathrm{E}[W_j]$) the closer $\mathrm{Var}(\epsilon_{G^\ast})$ is to zero.
Fig.~\ref{fig:Wdistr}, shows the weights' distribution for four different examples.
In Fig.~\ref{fig:Wdistr}, the neural networks and the respective filters and layers, were randomly selected out of the neural networks we consider in Section~\ref{sec:experimental}.
Similar results are obtained for the rest filters and neural networks.
As shown in Fig.~\ref{fig:Wdistr}, for all the examined filters, the majority of the weights is well concentrated in a closed region (squeezed dispersion in Fig.~\ref{fig:Wdistr}).
Hence, this feature boosts the efficiency of our variance reduction method, as explained above.

Using the $C$ value, obtained in~\eqref{eq:minvar}, that minimizes the variance, we compute the the mean convolution error $\mathrm{E}[\epsilon_{G^\ast}]$:
\begin{small}
\begin{equation}\label{eq:axconvmean}
\begin{split}
\mathrm{E}[\epsilon_{G^\ast}] &= \sum_{j=1}^{k}{\mathrm{E}\Big[\epsilon_j - v_j \Big]}\\
                     &= \sum_{j=1}^{k}{\mathrm{E}[x_j]\cdot W_j} - \sum_{j=1}^{k}{\mathrm{E}[x_j]\cdot \mathrm{E}[W_j]} \\
                     &= \underbrace{\frac{(2^m-1)}{2}}_{\mathrm{E}[x_j]}\Big(\sum_{j=1}^{k}{W_j}- k \cdot \mathrm{E}[W_j]\Big)\\
                     &= 0
\end{split}
\end{equation} 
\end{small}\noindent
As a result, the proposed control variate approximation method with $V=\mathrm{E}[W_j]\sum_{j=1}^{k}{x_j}$, effectively nullifies the mean error of the approximate convolution and also manages to decrease its variance.
In other words, the error distribution is constrained in a squeezed region around zero.
Hence, high convolution accuracy is expected.
However, as~\eqref{eq:var} shows, the larger the $m$ is, the larger the error variance will be and thus the accuracy loss.

\section{Approximate DNN Accelerator Implementation}\label{sec:architecture}
\begin{figure}[t!]
\centering
\resizebox{0.95\columnwidth}{!}{\includegraphics{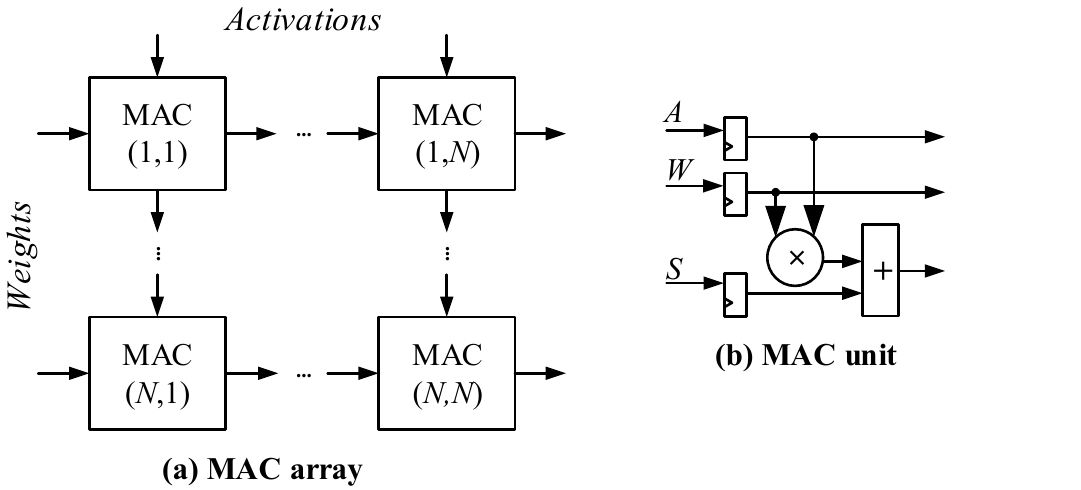}}
\caption{The a) accurate systolic MAC array and b) MAC unit.}
\label{fig:acmacarray}
\end{figure}

\begin{figure}[t!]
\centering
\resizebox{0.95\columnwidth}{!}{\includegraphics{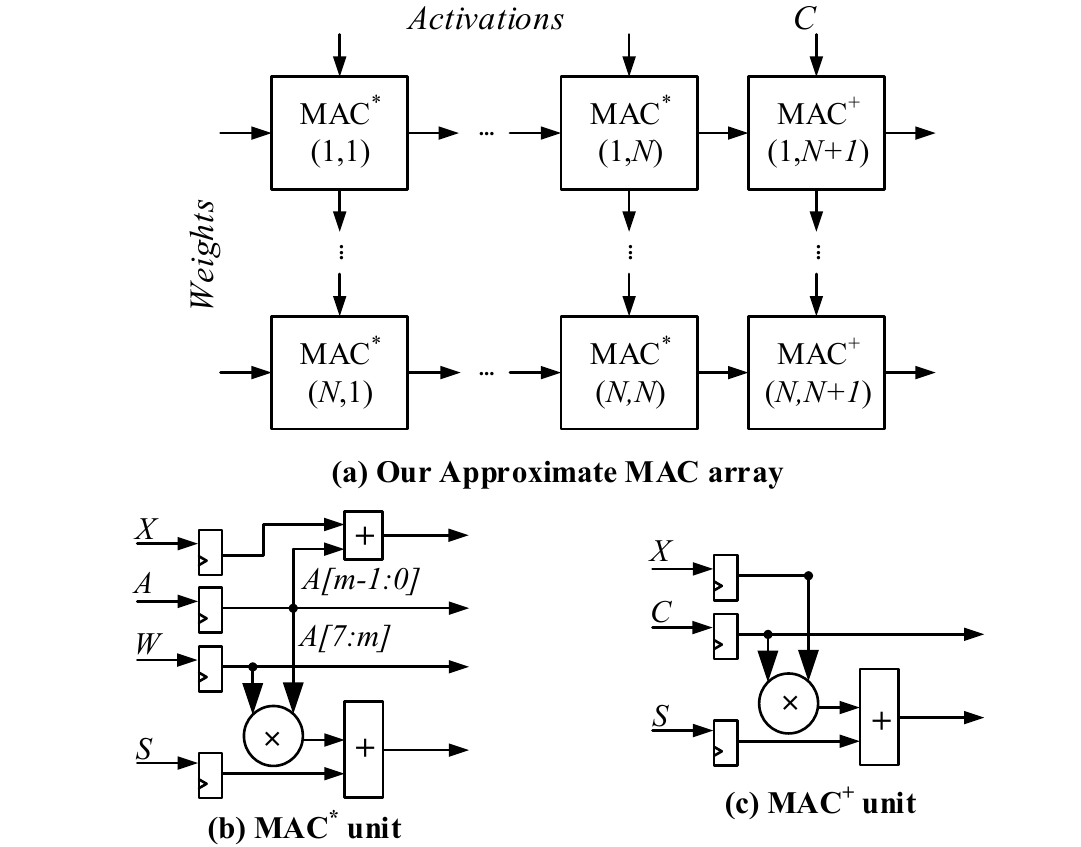}}
\caption{a) Our approximate systolic MAC array, b) the MAC$^*$ unit, and c) the MAC$^+$ unit.}
\label{fig:axmacarray}
\end{figure}

In this work, as our DNN accelerator use case, we consider a micro-architecture similar to the Google TPU~\cite{jouppi2017datacenter}.
TPU comprises a large $N\times N$ systolic MAC array.
Fig.~\ref{fig:acmacarray} illustrates the accurate systolic MAC array.
Fig.~\ref{fig:axmacarray} depicts how the accurate MAC array (Fig.~\ref{fig:acmacarray}) is modified to implement our proposed control variate approximation.
As shown in Fig.~\ref{fig:axmacarray}, our approximate MAC array features $N$ rows but $N+1$ columns.
In the first $N$ columns, MAC$^*$ units are used while the $N+1$ column uses MAC$^+$ units.
The first $N$ columns with the MAC$^*$ units calculate the convolution result $B+\sum_{j=1}^{k}{W_j\cdot A_j|_{approx}}$.
The MAC$^+$ unit in the last column is responsible for adding the control variable $V=\mathrm{E}[W_j]\sum_{j=1}^{k}{x_j}$ to the partial sum of the first $N$ columns.

In the accurate MAC array, each MAC unit comprises an $8$-bit multiplier and a $\lceil \log_2( N \times (2^{16}-1))\rceil$-bit adder to avoid accumulation overflow~\cite{tasoulas2020weight}.
For a $64\times 64$ MAC array, the size of the adder is 22-bit.

In the approximate MAC array, the MAC$^*$ unit uses the perforated multipliers~\cite{ZervakisTVLSI2016} to perform the approximate multiplication $W_j\cdot A_j|_{approx}$.
The approximate product requires $16-m$ bits, where $m$ is the number of perforated partial products.
Hence, we can also decrease the size of the adder by $m$ bits.
To achieve this, we have to shift the final partial sum $m$ places left, as we will explain later.
Moreover, for the partial sum input of the first column MAC$^*$, instead of $B$ we use $B[7:m]$, in order to align the two inputs of the adder.
In addition, each MAC$^*$ has to compute the partial sum required to calculate $V$ (i.e., $\sum_{j=1}^{k}{x_j}$).
As discussed in the previous section, the size of $x_j$ is $m$-bit and thus, a $\lceil \log_2( N \times (2^m-1))\rceil$-bit adder is required.
Note, however, that $m<8$.
For a $64\times 64$ MAC array and $m=2$, the size of the adder is only 8 bits.

Overall, each MAC$^*$ computes the following:
\begin{small}
\begin{equation}\label{eq:macstar}
\begin{gathered}
    P_j^*=W_j\cdot A_j[7:m]\\
    sum_j=sum_{j-1}+P_j^*\,,\ sum_0=B[7:m]\\
    sumX_j=sumX_{j-1}+A_j[m-1:0]\,,\ sumX_0=0
\end{gathered}
\end{equation}
\end{small}\noindent
In terms of hardware requirements, the multiplication requires the accumulation of $m$ fewer partial products and thus, $8\cdot m$ fewer full adders are required~\cite{LeonTVLSI2018}.
In addition, the adder that computes $sum_j$ requires $m$ fewer full adders, while the adder that computes $sumX_j$ requires $\lceil \log_2( N \times (2^m-1))\rceil-1$ full adders and $1$ half adder~\cite{LeonTVLSI2018}.
Hence, the full adders reduction is about ($9\cdot m$-$\lceil \log_2( N \times (2^m-1))\rceil+0.5$).

Moreover, $sumX_j$ and $sum_j$ are independent and are computed in parallel.
Hence, the adder $sumX_j$ is not on the critical path of MAC$^*$ and thus, a slow ripple-carry adder can be used to save power.
Furthermore, MAC$^*$ can operate at lower delay than the accurate MAC, since the delay of both the multiplier and the $sum_j$ adder is decreased.
The multiplier has to accumulate $m$ fewer partial products (i.e., shorter accumulation tree~\cite{ZervakisTVLSI2016}), while the adder has to sum $m$ fewer bits.
Therefore, this delay slack enables downsizing of gates of the critical
paths and boosts further the area and power savings~\cite{VenkataramaniDATE2013}.

As shown in Fig.~\ref{fig:axmacarray}, the last column of each row contains a MAC$^+$ unit.
MAC$^+$ computes the following:
\begin{small}
\begin{align}
    V=C\cdot sumX_{N}\,,\ \text{where}\ C=\mathrm{E}[W_j] \label{eq:macplusP} \\
    G^\ast=\{sum_N,B[m-1:0]\}+V \label{eq:macplusS} 
\end{align}
\end{small}
MAC$^+$ requires an accurate $\lceil \log_2( N \times (2^m-1))\rceil \times 8$-bit multiplier to compute the product of \eqref{eq:macplusP}, i.e., $V=E[W_j]\sum_{j=1}^{k}{x_j}$.
Furthermore, a $\lceil \log_2( N \times (2^{16}-1))\rceil$-bit adder is required to produce the final output.
The length of this adder and the length of the adder required by the accurate MAC unit are the same.
Note that in~\eqref{eq:macplusS}, the partial sum $sum_N$, i.e., $\sum_{j=1}^{k}{W_j\cdot A_j|_{approx}}$, is shifted left $m$ places and in its $m$-LSBs the missing $m$-LSBs of B, i.e., $B[m-1:0]$, are added.
As a result, we achieve to both i) shift left to the required position the partial sum and ii) add the missing bits of the bias.

If the delay of MAC$^+$ is higher than the delay of the accurate MAC, we pipeline MAC$^+$ in order to sustain the same operating frequency.
Hence, if the MAC$^+$ unit requires $l$ cycles, then the latency overhead of our proposed approximate MAC array is $l$ cycles per convolution layer.
However, considering that the inference phase requires thousands of cycles for each convolution layer~\cite{samajdar2018scale}, this overhead is negligible.
Moreover, the MAC$^+$ requires the value $C=\mathrm{E}[W_j]$ to calculate $V$.
This value can be transferred to the DNN accelerator among with the filters' weights.
The memory overhead of the latter, is only 8 bits per filter.

\begin{table}[t!]
\renewcommand{\arraystretch}{1.2}
\caption{Theoretical Evaluation of Full Adders (FA) Reduction}
\label{tab:fa}
\footnotesize
\centering
  \begin{tabular}[t]{c|c|c|c}
  \hline
$\boldsymbol{N}$ & \makecell{\textbf{FA decrease} \\ \textbf{due to MAC}$^\mathbf{*}$} & \makecell{\textbf{FA Increase} \\ \textbf{due to MAC}$^\mathbf{+}$} & \makecell{\textbf{Total FA} \\ \textbf{Decrease}} \\ \hline
\multicolumn{4}{c}{$m=1$} \\ \hline
16 & 1408 & 760 & 648 \\ \hline
32 & 4608 & 1776 & 2832 \\ \hline
48 & 8064 & 3048 & 5016 \\ \hline
64 & 14336 & 4064 & 10272 \\ \hline
\multicolumn{4}{c}{$m=2$} \\ \hline
16 & 3200 & 984 & 2216 \\ \hline
32 & 11776 & 2224 & 9552 \\ \hline
48 & 24192 & 3720 & 20472 \\ \hline
64 & 43008 & 4960 & 38048 \\ \hline
  \end{tabular}
\end{table}

As aforementioned, our approach replaces the accurate MAC units with the MAC$^*$ ones.
Hence, the power and area of $N\times N$ MAC units are reduced.
However, $N$ additional MAC$^+$ units are required to add $V$.
The impact of MAC$^+$ becomes negligible as the size of the MAC array increases.
The gains due to MAC$^*$ increase quadratically w.r.t. $N$ while the overhead due to MAC$^+$ increases linearly w.r.t. $N$.
Table~\ref{tab:fa} presents a theoretical estimation of the full adders reduction for perforation values $m=1$ and $m=2$.
In addition, Table~\ref{tab:fa} shows both the reduction due to MAC$^*$ and the overhead due to the additional additional MAC$^+$.
The number of required full adders is calculated based on~\cite{LeonTVLSI2018}.
As shown in Table~\ref{tab:fa}, as $N$ (MAC array size) or $m$ (the perforation parameter) increase, the number of full adders required, decreases significantly.
In addition, as $N$ increases, the reduction due to MAC$^*$ becomes significantly larger than the increase due to MAC$^+$.
As a result, the impact of MAC$^+$ is insignificant compared to the gains due MAC$^*$.
Even for small MAC arrays, e.g., $N=16$, when $m=1$, the reduction due to MAC$^*$ is $2.59$x higher than the overhead due to MAC$^+$.
Similarly, when $m=2$, the respective value is $3.25$x.
Therefore, Table~\ref{tab:fa} validates our hypothesis in Section~\ref{sec:error} that the hardware cost of our proposed control variate method is negligible.

Table~\ref{tab:fa} provides a simplistic theoretical evaluation of the proposed architecture in order to provide an initial insight on the impact of the MAC$^*$ and MAC$^+$ units.
For this reason, we use the gains in full adders as a representative metric.
Components such as registers and partial product generation circuits are not considered.
In addition, the impact of logic downsizing is not evaluated either.
Section~\ref{sec:experimental} provides a comprehensive experimental evaluation of the hardware gains.

\section{Experimental Results}\label{sec:experimental}
In this section, we examine the efficiency our proposed control variate approximation in terms of area, power, and accuracy.
For our analysis we employ industry-strength tools.
All the MAC arrays evaluated in this section, are synthesized using Synopsys Design Compiler and are mapped to a commercial $14$nm technology library.
Synthesis is performed using the ``compile\_ultra'' command and targeting performance optimization.
Each accurate MAC array is synthesized at its critical path delay, while the approximate ones are synthesized at the critical path delay of the respective accurate one.
Hence, the accurate and the respective approximate MAC arrays feature the same clock frequency.
Post-synthesis timing simulations are performed using Mentor Questasim to obtain precise switching activity.
Then, Synopsys PrimeTime is used to calculate the power consumption.
For power analysis, we simulate the examined MAC arrays for 10,000 inference cycles to obtain accurate switching activity estimation.
Running post-synthesis timing simulations for the entire inference phase is infeasible due to the vast time required~\cite{tasoulas2020weight}.
The inference accuracy is captured by extending the approximate TensorFlow implementation of~\cite{tfapproxDate2020} with our control variate approximation.
Six NNs of varying size, depth, and architecture are considered.
The NNs are trained on the Cifar-10 and Cifar-100 datasets.

\subsection{Hardware Evaluation}
First, we evaluate the power and area gains of our proposed approximate MAC array compared to the accurate one.
Both MAC aarays are implemented using the optimized arithmetic components of the Synopsys DesignWare Library, as typically done in commercial flows.
Note that, unlike logic approximation approaches~\cite{alwann,zervakis2020design,tasoulas2020weight}, the perforated multiplier, we use, applies functional approximation and thus, it can be implemented with the optimized DesignWare components~\cite{SaadatDATE2020,ZervakisTVLSI2016}.
Finally, note that for the examined sizes, i.e., $16\times 16$ to $64\times 64$, the MAC$^+$ unit required $1$ cycle and thus, the overall overhead is only $1$ cycle per convolution layer.

\begin{figure}[t!]
\centering
\resizebox{0.95\columnwidth}{!}{\includegraphics{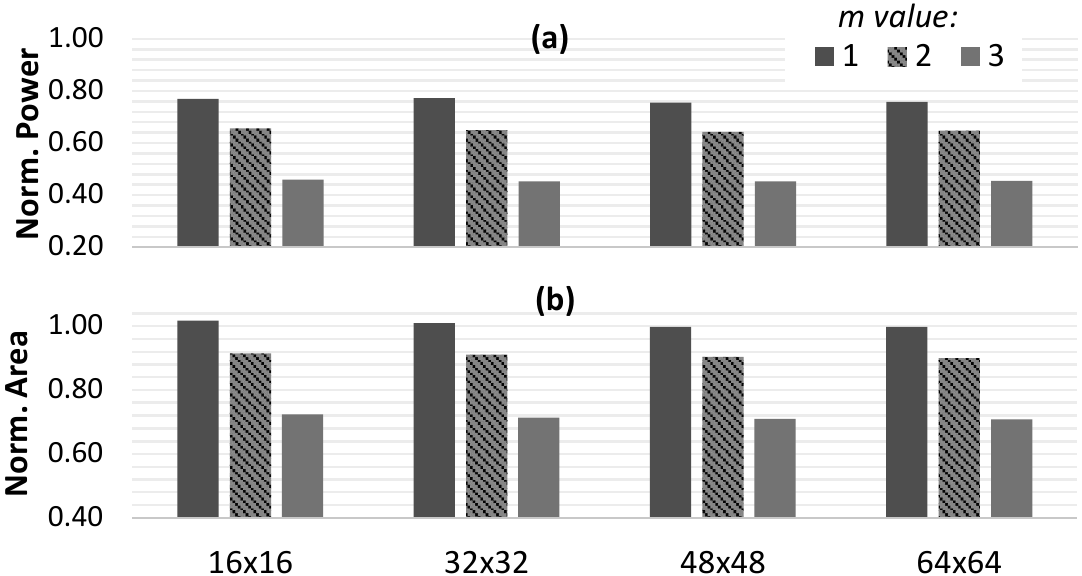}}
\caption{The a) power and b) area of our control variate approximation for varying $m$ values and MAC array sizes. The area and power values are normalized over the corresponding values of the respective accurate design.}
\label{fig:hw}
\end{figure}

Fig.~\ref{fig:hw} presents the area and power savings delivered by our technique for varying MAC array sizes and perforation values (i.e., $m$).
As shown in Fig.~\ref{fig:hw}a, our approximate MAC arrays achieve large power reduction, ranging form $22.8$\% up to $54.7$\%.
In addition, it is observed that the power gain depends mainly on the perforation value $m$.
For $m=1$, the power reduction ranges from $22.8$\% to $24.2$\%.
For $m=2$, the respective values are from $34.5$\% to $35.7$\%, while for $m=3$, the power decreases from $54.1$\% to $54.8$\%.
This behavior verifies that the impact of the MAC$^+$ units is negligible and thus, the power reduction of the MAC array is mainly defined by the power reduction delivered by the MAC$^*$ units.
In Fig.~\ref{fig:hw}b, the area reduction follows the same trend, i.e., it is defined by the $m$ value and is slightly affected by the array size.
The area gain goes up to $29$\% for $m$=3.
For $m=1$, the area gains due to the reduction in combinational logic is compensated by the increase in the number of registers required in the MAC$^*$ units (compared to the accurate MAC unit).
Hence, for $m=1$ the area remains almost the same.

In Table~\ref{tab:macplus}, we evaluate the impact of the MAC$^+$ units that are required to add $V$.
Table~\ref{tab:macplus} presents the area and power of the MAC$^+$ units w.r.t. the area and power of the entire approximate MAC array.
For all the examined scenarios the MAC$^+$ units occupy at most the $1.49$\% of the entire array.
Similarly, the power consumed by the MAC$^+$ units is up to $1.87$\% of the total power consumption.
It is noteworthy, that these values correspond to the smallest MAC array (only $16\times 16$) and the highest approximation ($m=3$).
As shown in Table~\ref{tab:macplus}, as the MAC array size increases, the impact of the MAC$^+$ units becomes insignificant, validating our claim in Section~\ref{sec:architecture}.
For example, the respective values for the $64\times 64$ MAC array are only $0.40$\% and $0.49$\%.
Finally, Fig.~\ref{fig:hw} and Table~\ref{tab:macplus} demonstrate the scalability of our technique/architecture, i.e., the achieved power savings are sustained independently of the MAC array size.

\begin{table}[t!]
\renewcommand{\arraystretch}{1.2}
\caption{Evaluation of the Area and Power Overheads of MAC$^+$}
\label{tab:macplus}
  \footnotesize
\centering
  \begin{tabular}[t]{c|c|c|c|c}
 \hline
  \multicolumn{5}{c}{\textbf{Percentage of Total Area (\%)}} \\ \hline
$m$ & $16\times 16$ & $32\times 32$ & $48\times 48$ & $64\times 64$ \\ \hline
1 & 1.06 & 0.55 & 0.38 & 0.28\\ \hline
2 & 1.18 & 0.61 & 0.41 & 0.31\\ \hline
3 & 1.49 & 0.77 & 0.53 & 0.40\\ \hline
  \multicolumn{5}{c}{\textbf{Percentage of Total Power (\%)}} \\ \hline
$m$ & $16\times 16$ & $32\times 32$ & $48\times 48$ & $64\times 64$ \\ \hline
1 & 1.15 & 0.58 & 0.40 & 0.30\\ \hline
2 & 1.32 & 0.68 & 0.46 & 0.35\\ \hline
3 & 1.87 & 0.97 & 0.64 & 0.49\\ \hline
  \end{tabular}
\vspace{10pt}  
\caption{Accuracy Evaluation Over Six Neural Networks Trained on Cifar-10 and Cifar-100}
\label{tab:acc}
\footnotesize
\centering
\begin{threeparttable}
  \begin{tabular}[t]{l|c|c|c|c|c|c}
  \hline
\multicolumn{7}{c}{\textbf{Accuracy Loss (\%)}} \\ \hline
\multirow{2}{*}{\makecell{\textbf{NN on}\\ \textbf{Cifar-10}}}  & \multicolumn{2}{c|}{$m=1$} & \multicolumn{2}{c|}{$m=2$} & \multicolumn{2}{c}{$m=3$} \\ \cline{2-7}
  & \textbf{Ours}\tnote{+} & \textbf{ w/o $\boldsymbol V$}\tnote{*} & \textbf{Ours} & \textbf{ w/o $\boldsymbol V$} & \textbf{Ours} & \textbf{ w/o $\boldsymbol V$} \\ \hline
googlenet & -0.16 & 0.35 & 0.00 & 4.13 & 1.95 & 31.78\\ \hline
ResNet44 & 0.03 & 0.73 & 0.83 & 4.49 & 5.75 & 32.94\\ \hline
ResNet56 & 0.49 & 0.60 & 1.25 & 5.36 & 5.94 & 39.01\\ \hline
shufflenet & 0.12 & 3.40 & -0.48 & 7.6 & 6.53 & 31.74\\ \hline
vgg13 & -0.01 & 0.23 & -0.30 & 0.76 & 0.69 & 6.76\\ \hline
vgg16 & -0.13 & 0.19 & 0.38 & 1.66 & 3.86 & 8.71\\ \hline
\textbf{Average} & 0.06 & 0.92 & 0.28 & 4.00 & 4.12 & 25.16\\ \hline \hline
\multirow{2}{*}{\makecell{\textbf{NN on}\\ \textbf{Cifar-100}}}    & \multicolumn{2}{c|}{$m=1$} & \multicolumn{2}{c|}{$m=2$} & \multicolumn{2}{c}{$m=3$} \\ \cline{2-7}
  & \textbf{Ours} & \textbf{ w/o $\boldsymbol V$} & \textbf{Ours} & \textbf{ w/o $\boldsymbol V$} & \textbf{Ours} & \textbf{ w/o $\boldsymbol V$} \\ \hline
googlenet & 0.05 & 2.43 & 1.47 & 13.19 & 7.02 & 44.52\\ \hline
ResNet44 & 0.77 & 1.02 & 1.82 & 14.17 & 11.27 & 43.4\\ \hline
ResNet56 & -0.34 & 3.02 & 2.25 & 15.83 & 14.34 & 44.59\\ \hline
shufflenet & 0.20 & 9.47 & 1.09 & 5.92 & 6.57 & 15.84\\ \hline
vgg13 & 0.89 & 3.01 & 1.91 & 5.89 & 3.98 & 14.7\\ \hline
vgg16 & -0.03 & 3.96 & 0.03 & 2.41 & 2.8 & 11.54\\ \hline
\textbf{Average} & 0.26 & 3.82 & 1.43 & 9.57 & 7.92 & 29.10\\ \hline
  \end{tabular}
\begin{tablenotes}\footnotesize
\item[+] Accuracy achieved when using the approximate multiplier of~\cite{ZervakisTVLSI2016} with our proposed control-variate approximation.
\item[*] Accuracy achieved when using only the approximate multiplier of~\cite{ZervakisTVLSI2016}.
\end{tablenotes}
\end{threeparttable}
\end{table}

\subsection{Accuracy Evaluation}
Next, we evaluate the accuracy delivered by our control variate approximation.
For our analysis, we consider six NNs and in Table~\ref{tab:acc} we report the accuracy loss for varying perforation values $m$.
Note that the accuracy depends only on $m$ and not on the size of the MAC array.
Moreover, in order to analyze the efficiency of our control variate approximation, we report the accuracy when using the same perforated multipliers~\cite{ZervakisTVLSI2016} (same $m$) without our proposed control variate technique (i.e.,~\cite{ZervakisTVLSI2016} w/o $V$).
Negative values in Table~\ref{tab:acc} refer to accuracy improvement due to the use of approximation~\cite{alwann}.
As shown in Table~\ref{tab:acc}, the average accuracy loss of our method for Cifar-10 is $0.06$\%, $0.28$\%, and $4.12$\% for $m=1$, $m=2$, and $m=3$, respectively.
The corresponding values for the more challenging Cifar-100 dataset are $0.26$\%, $1.43$\%, and $7.92$\%.
As a result, our technique achieves \mytilde$24$\% power reduction for negligible accuracy loss, i.e., $0.16$\% on average on both datasets for $m=1$.
The power gains rise to \mytilde$35$\% ($m=2$) for an average accuracy loss of only $0.85$\%.
Finally, for $6.02$\% average accuracy loss ($m=3$), the power savings jump to \mytilde$55$\%.

Table~\ref{tab:acc} highlights also the efficiency of our control variate approximation in decreasing the convolution error.
Compared with~\cite{ZervakisTVLSI2016} w/o $V$, i.e., same approximation without our control variate, our technique achieves $2$\%, $6$\%, and $21$\% higher accuracy, on average, for $m=1$, $m=2$, and $m=3$, respectively.
As a result, considering Tables~\ref{tab:macplus}-\ref{tab:acc}, our proposed control variate approximation delivers significant accuracy improvement for minimal hardware cost.

\subsection{Comparison with State of the Art}
Finally, we compare our control variate approximation against three recent state-of-the-art works~\cite{alwann,zervakis2020design,tasoulas2020weight} that similar to our approach do not require retraining.
In~\cite{alwann} a non-uniform architecture is used, that contains several approximate multiplier types from~\cite{mrazek2017evoapproxsb}.
Each convolution layer uses only one multiplier type and the rest ones are power-gated.
However, this leads to high area overhead and throughput loss due to the power-gated multipliers.
For the fairness of the evaluation, we consider a uniform architecture for~\cite{alwann}, with only one approximate multiplier type.
In addition, note that the non-uniform approach of~\cite{alwann} is orthogonal with our work, since our technique also supports instantiating several MAC arrays with varying $m$ values.
For our comparison, we consider a $64\times 64$ MAC array.

\begin{figure}[t!]
\centering
\resizebox{0.90\columnwidth}{!}{\includegraphics{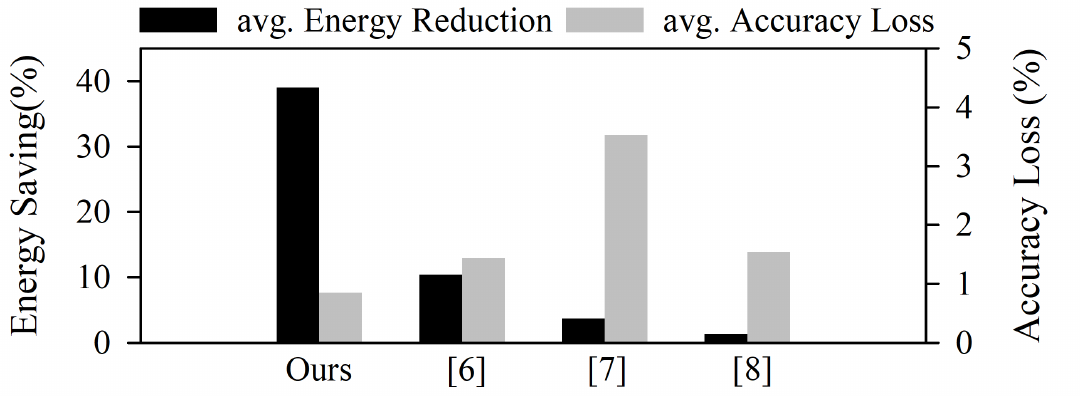}}
\caption{Comparison of our technique against the state of the art~\cite{alwann,zervakis2020design,tasoulas2020weight}. The average energy reduction and average accuracy loss w.r.t. the NNs examined in Table~\ref{tab:acc} is reported.}
\label{fig:soa}
\end{figure}

The architectures of~\cite{alwann,zervakis2020design,tasoulas2020weight} are based on the multipliers of EvoApprox library~\cite{mrazek2017evoapproxsb}.
On the other hand, as typically done in commercial design flows, we employed the industry-level DesignWare library for our implementation.
We implemented $64\times 64$ MAC arrays using~\cite{alwann,zervakis2020design,tasoulas2020weight} and we evaluated their hardware characteristics.
However, since~\cite{alwann,zervakis2020design,tasoulas2020weight} are based on suboptimal mutlipliers, the obtained MAC arrays feature significantly higher power, area, and delay compared with our \textit{accurate} MAC array implemented with the DesignWare components.
For this reason, we use also EvoApprox library~\cite{mrazek2017evoapproxsb} as the base of the MAC array implemented with our technique.
For our technique, we select the perforation value $m=2$ since, as shown above, it features high power reduction for moderate accuracy loss.

To compare our work against the state-of-the-art works~\cite{alwann,zervakis2020design,tasoulas2020weight}, we evaluate, for each technique, the energy consumption and accuracy loss for all the NNs in Table~\ref{tab:acc} (for both Cifar-10 and Cifar-100).
Energy consumption is calculated by $cycles\times power \times delay$.
The cycles are obtain using the CNN cycle accurate simulator of ARM~\cite{samajdar2018scale}.
We evaluate the energy consumption since i) our technique requires $1$ additional cycle per convolution layer and ii)~\cite{zervakis2020design,tasoulas2020weight} use approximate reconfigurable multipliers with different configuration for each convolution layer.
Fig.~\ref{fig:soa} reports, for each technique, the average energy reduction and accuracy loss.
The energy reduction and accuracy loss are reported with respect to the corresponding values of the accurate design.
The accurate MAC array is also implemented using the accurate multiplier (1JFF) of~\cite{mrazek2017evoapproxsb}.
For~\cite{alwann}, the results for the 125K multiplier are presented, since it delivered the the best energy-accuracy tradeof among all the approximate multipliers of~\cite{mrazek2017evoapproxsb}.
As shown in Fig.~\ref{fig:soa}, our technique delivers the highest energy savings.
Compared with~\cite{tasoulas2020weight},~\cite{alwann}, and~\cite{zervakis2020design} our technique achieves $3.8$x, $10.5$x, and $29.2$x higher average energy reduction, respectively.
In addition, as Fig.~\ref{fig:soa} shows, all the techniques achieve comparable accuracy, with our technique being the most accurate.
However, in order to sustain high accuracy,~\cite{alwann,zervakis2020design,tasoulas2020weight} apply conservative approximation.
As a result,~\cite{alwann,zervakis2020design,tasoulas2020weight} achieve very limited energy gains.
On the other hand, our control variate approximation reduces significantly the error and enables, thus, using highly approximate multipliers, such as~\cite{ZervakisTVLSI2016}. 
Therefore, our control variate approximation boosts the obtained gains for minimal loss in accuracy.

\section{Conclusion}
In this work, we introduced control variate approximation to increase the accuracy of approximate DNN accelerators without requiring any DNN retraining.
Our mathematical analysis demonstrates that our technique mitigates the error induced by the approximate multipliers, by effectively nullifying the mean convolution error and reducing its variance. 
Our control variate approximation enables using aggressive approximate multipliers to design approximate DNN accelerators that boost the power savings.

\section*{Acknowledgment}
This work is partially supported by the German Research Foundation (DFG) through the project ``ACCROSS: Approximate Computing aCROss the System Stack''.

\end{document}